\pdfoutput=1
\documentclass[runningheads,envcountsame, a4paper]{llncs}
\usepackage[misc]{ifsym}
\usepackage{soul}
\usepackage{url}
\usepackage[utf8]{inputenc}
\usepackage{graphicx}
\usepackage{amsmath}
\usepackage{amsfonts}
\usepackage{booktabs}
\usepackage{caption}
\usepackage{xspace} 
\usepackage{multirow}
\usepackage{diagbox}
\usepackage{subfigure}
\usepackage{braket}
\usepackage[linesnumbered,ruled,vlined]{algorithm2e}

\newcommand{\dataSet}{\textsl}
\newcommand{\ringroad}{\dataSet{$Ring$}\xspace}
\newcommand{\intersectionone}{\dataSet{$Intersection$}\xspace}

\newcommand{\hangzhoufour}{\dataSet{$HZ_{4\times 4}$}\xspace}
\newcommand{\LAfour}{\dataSet{$LA_{1\times4}$}\xspace}

\newcommand{\indnet}{\dataSet{InDNet}\xspace}

\newcommand{\methodFont}{\textsl}
\newcommand{\learntosim}{\methodFont{ImIn-GAIL}\xspace}
\newcommand{\gail}{\methodFont{GAIL}\xspace}
\newcommand{\bc}{\methodFont{BC}\xspace}
\newcommand{\cfmrs}{\methodFont{CFM-RS}\xspace}
\newcommand{\cfmts}{\methodFont{CFM-TS}\xspace}

\newcommand{\problemSetFont}{\mathcal}
\newcommand{\traj}{\problemSetFont{\tau}}
\newcommand{\Traj}{\problemSetFont{T}}
\newcommand{\TrajD}{\problemSetFont{T}^D}
\newcommand{\policy}{\problemSetFont{\pi}}

\newcommand{\Expect}{\mathbb{E}}
\newcommand{\Real}{\mathbb{R}}
\newcommand{\Loss}{\mathcal{L}}

\newcommand{\DiscT}{\mathcal{D}_\psi}

\begin{document}
\title{Learning to Simulate on Sparse Trajectory Data}
%
%
\author{Hua Wei\inst{1}\Letter \and Chacha Chen\inst{1} \and Chang Liu\inst{2} \and Guanjie Zheng\inst{1} \and Zhenhui Li\inst{1}}
\authorrunning{H. Wei et al.}
%
\institute{Pennsylvania State University, University Park, PA 16802, USA \\
\email{\{hzw77, gjz5038, jessieli\}@ist.psu.edu, cjc6647@psu.edu} \\
\and
Shanghai Jiao Tong University, Shanghai, China \\
\email{only-changer@sjtu.edu.cn}}
\maketitle              
\begin{abstract}
Simulation of the real-world traffic can be used to help validate the transportation policies. A good simulator means the simulated traffic is similar to real-world traffic, which often requires dense traffic trajectories (i.e., with high sampling rate) to cover dynamic situations in the real world.
However, in most cases, the real-world trajectories are sparse, which makes simulation challenging. In this paper, we present a novel framework \learntosim to address the problem of learning to simulate the driving behavior from sparse real-world data. The proposed architecture incorporates data interpolation with the behavior learning process of imitation learning. To the best of our knowledge, we are the first to tackle the data sparsity issue for behavior learning problems. We investigate our framework on both synthetic and real-world trajectory datasets of driving vehicles, showing that our method outperforms various baselines and state-of-the-art methods. 

\end{abstract}

\section{Introduction}

Simulation of the real world is one of the feasible ways to verify driving policies on autonomous vehicles and transportation policies like traffic signal control ~\cite{wei2019presslight,wei2019colight,hua2018intellilight} or speed limit setting~\cite{wu2018differential} since it is costly to validate them in the real world directly~\cite{wei2019survey}. The driving behavior model, i.e., how the vehicle accelerates/decelerates, is the critical component that affects the similarity of the simulated traffic to the real-world traffic~\cite{krauss1998microscopic,leutzbach1986development,nagel1992cellular}. Traditional methods to learn the driving behavior model usually first assumes that the behavior of the vehicle is only influenced by a small number of factors with predefined rule-based relations, and then calibrates the model by finding the parameters that best fit the observed data~\cite{kesting2008calibrating,osorio2019efficient}. The problem with such methods is that their assumptions oversimplify the driving behavior, resulting in the simulated driving behavior far from the real world.
 
In contrast, imitation learning (IL) does not assume the underlying form of the driving behavior model and directly learns from the observed data (also called demonstrations from expert policy in IL literature). With IL, a more sophisticated driving behavior policy can be represented by a parameterized model like neural nets and provides a promising way to learn the models that behave similarly to expert policy.
Existing IL methods (e.g., behavior cloning~\cite{michie1990cognitive,torabi2018behavioral} and generative adversarial imitation learning~\cite{ho2016generative,bhattacharyya2018multi,song2018multi,zheng2020learing2sim}) for learning driving behavior relies on a large amount of behavior trajectory data that consists of dense vehicle driving states, either from vehicles installed with sensors, or roadside cameras that capture the whole traffic situation (including every vehicle driving behavior at every moment) in the road network.

\begin{figure}[t!]
  \centering
  \includegraphics[width=.75\linewidth]{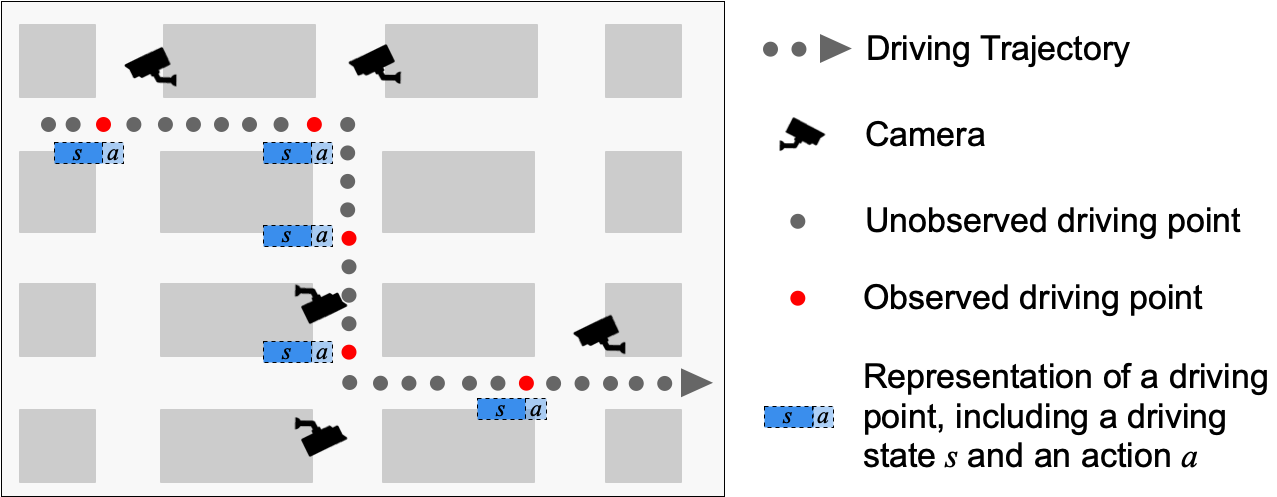}
  \caption{Illustration of a driving trajectory. In the real-world scenario, only part of the driving points can be observed and form a sparse driving trajectory (in red dots). Each driving point includes a driving state and an action of the vehicle at the observed time step. Best viewed in color.}
  \label{fig:intro}
\end{figure}

However, in most real-world cases, the available behavior trajectory data is sparse, i.e., the driving behavior of the vehicles at every moment is difficult to observe. It is infeasible to install sensors for every vehicle in the road network or to install cameras that cover every location in the road network to capture the whole traffic situation. Most real-world cases are that only a minimal number of cars on the road are accessible with dense trajectory, and the driving behavior of vehicles can only be captured when the vehicles drive near the locations where the cameras are installed. For example, in Figure~\ref{fig:intro}, as the cameras are installed only around certain intersections, consecutive observed points of the same car may have a large time difference, resulting in a sparse driving trajectory. 
As data sparsity is considered as a critical issue for unsatisfactory accuracy in machine learning, directly using sparse trajectories to learn the driving behavior could make the model fail to learn the behavior policy at the unobserved states. 

To deal with sparse trajectories, a typical approach is to interpolate the sparse trajectories first and then learn the model with the dense trajectories~\cite{li2016scalable,yi2016st,zheng2012reducing}. This two-step approach also has an obvious weakness, especially in the problem of learning behavior models. 
For example, linear interpolation is often used to interpolate
the missing points between two observed trajectory points. But in real-world cases, considering the interactions between vehicles, the vehicle is unlikely to drive at a uniform speed during that unobserved time period, hence the interpolated trajectories may be different from the true trajectories. 
However, the true trajectories are also unknown and are exactly what we aim to imitate. A better approach is to integrate interpolation with imitation because they should inherently be the same model. To the best of our knowledge, none of the existing literature has studied the real-world problem of learning driving policies from sparse trajectory data.

In this paper, we present \learntosim, an approach that can learn the driving behavior of vehicles from observed sparse trajectory data. 
\learntosim learns to mimic expert behavior under the framework of generative adversarial imitation learning (GAIL), which learns a policy that can perform expert-like behaviors through rewarding the policy for deceiving a discriminator trained to classify between policy-generated and expert trajectories.
Specifically, for the data sparsity issue, we present an interpolator-discriminator network that can perform both the interpolation and discrimination tasks, and a downsampler that draws supervision on the interpolation task from the trajectories generated by the learned policy. We conduct experiments on both synthetic and real-world data, showing that our method can not only have excellent imitation performance on the sparse trajectories but also have better interpolation results compared with state-of-the-art baselines. 
The main contributions of this paper are summarized as follows:
\begin{itemize}
    \item We propose a novel framework \learntosim, which can learn driving behaviors from the real-world sparse trajectory data.
    \item We naturally integrate the interpolation with imitation learning that can interpolate the sparse driving trajectory.
    \item We conduct experiments on both real and synthetic data, showing that our approach significantly outperforms existing methods. We also have interesting cases to illustrate the effectiveness on the imitation and interpolation of our methods.
\end{itemize}

\section{Preliminaries}
\label{sec:prelim}

\begin{definition}[Driving Point]
A driving point $\traj^t=(s^t, a^t, t)$ describes the driving behavior of the vehicle at time $t$, which consists of a driving state $s^t$ and an action $a^t$ of the vehicle. Typically, the state $s^t$ describes the surrounding traffic conditions of the vehicle (e.g., speed of the vehicle and distance to the preceding vehicle), and the action $a^t\sim\policy(a|s^t)$ the vehicle takes at time $t$ is the magnitude of acceleration/deceleration following its driving policy $\policy(a|s^t)$.
\end{definition}

\begin{definition}[Driving Trajectory]
A driving trajectory of a vehicle is a sequence of driving points generated by the vehicle in geographical spaces, usually represented by a series of chronologically ordered points, e.g. $\traj = ( \traj^{t_0},\cdots ,\traj^{t_N} )$.
\end{definition}

In trajectory data mining~\cite{liu2020online,lou2009map,zheng2015trajectory}, a \emph{dense trajectory} of a vehicle is the driving trajectory with high-sampling rate (e.g., one point per second on average), and a \emph{sparse trajectory} of a vehicle is the driving trajectory with low-sampling rate (e.g., one point every 2 minutes on average). In this paper, the observed driving trajectory is a sequence of driving points with large and irregular intervals between their observation times.

\begin{problem}
In our problem, a vehicle observes state $s$ from the environment, take action $a$ following policy $\policy^E$ at every time interval $\Delta t$, and generate a raw driving trajectory $\traj$ during certain time period. While the raw driving trajectory is dense (i.e., at a high-sampling rate), in our problem we can only observe a set of sparse trajectories $\mathbb{\Traj}_E$ generated by expert policy $\policy^E$ as expert trajectory, where $\Traj_E = \{\traj_i |\traj_i = ( \traj_i^{t_0},\cdots ,\traj_i^{t_N} ) \}$, $t_{i+1}-t_{i} \gg \Delta t$ and $t_{i+1}-t_{i}$ may be different for different observation time $i$. Our goal is to learn a parameterized policy $\policy_\theta$ that imitates the expert policy $\policy^E$.
\end{problem}


\section{Method}
\label{sec:method}

In this section, we first introduce the basic imitation framework, upon which we propose our method (\learntosim) that integrates trajectory interpolation into the basic model.

\subsection{Basic GAIL Framework}
In this paper, we follow the framework similar to GAIL~\cite{ho2016generative} due to its scalability to the multi-agent scenario and previous success in learning human driver models~\cite{kuefler2017imitating}. GAIL formulates imitation learning as the problem of
learning policy to perform expert-like behavior by rewarding it for ``deceiving'' a classifier trained to discriminate between policy-generated and expert state-action pairs. For a neural network classifier $\DiscT$ parameterized by $\psi$, the GAIL objective
is given by $\mathop{max}_{\psi} \mathop{min}_{\theta} \Loss(\psi,\theta)$ where $\Loss(\psi,\theta)$ is :

\begin{equation}
\label{eq:gail}
\begin{aligned}
    \Loss(\psi,\theta) = \Expect_{(s,a)\sim \traj \in \Traj_E} \log\DiscT(s,a) + \Expect_{(s,a)\sim \traj \in \Traj_G} \log (1-\DiscT(s,a)) - \beta H(\policy_\theta)
\end{aligned}
\end{equation}
where $\Traj_E$ and $\Traj_G$ are respectively the expert trajectories and the generated trajectories from the interactions of policy $\policy_\theta$ with the simulation environment, $H(\policy_\theta)$ is an entropy regularization term.

~$\bullet$ \emph{Learning $\psi$}:
When training $\DiscT$, Equation~\eqref{eq:gail} can simply be set as a sigmoid cross entropy where positive samples are from $\Traj_E$ and negative samples are from $\Traj_G$. Then optimizing $\psi$ can be easily done with gradient ascent.

~$\bullet$ \emph{Learning $\theta$}:
The simulator is an integration of physical rules, control policies and randomness and thus its parameterization is assumed to be unknown. Therefore, given $\Traj_G$ generated by $\policy_\theta$ in the simulator, Equation~\eqref{eq:gail} is non-differentiable w.r.t $\theta$. In order to learn $\policy_\theta$, GAIL optimizes through reinforcement learning, with a surrogate reward function
formulated from Equation~\eqref{eq:gail} as:

\begin{equation}
\label{eq:gail-reward}
    \tilde{r}(s^t, a^t; \psi) = - \log(1 - \DiscT(s^t, a^t))
\end{equation}

Here, $\tilde{r}(s^t, a^t; \psi)$ can be perceived to be useful in driving $\policy_\theta$ into regions of the state-action space at time $t$ similar to those
explored by $\policy^E$. Intuitively, when the observed trajectory is dense, the surrogate reward from the discriminator in Equation~\eqref{eq:gail-reward} is helpful to learn the state transitions about observed trajectories. However, when the observed data is sparse, the reward from discriminator will only learn to correct the observed states and fail to model the behavior policy at the unobserved states. To relieve this problem, we propose to interpolate the sparse expert trajectory within the based imitation framework.

\subsection{Imitation with Interpolation}
\begin{figure*}[t!]
  \centering
  \includegraphics[width=.80\linewidth]{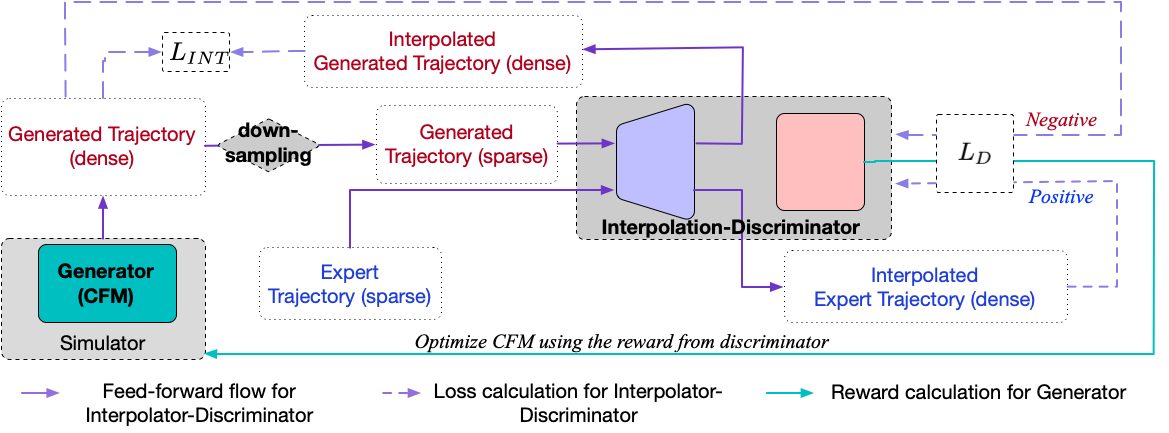}
  \caption{Proposed \learntosim Approach. The overall framework of \learntosim includes three components: generator, downsampler, and interpolation-discriminator. Best viewed in color.}
  \label{fig:framework}
\end{figure*}

An overview of our proposed Imitation-Interpolation framework (\learntosim) is shown in Figure~\ref{fig:framework}, which consists of the following three key components.

\subsubsection{Generator in the simulator}
Given an initialized driving policy $\policy_\theta$, the dense trajectories $\TrajD_G$ of vehicles can be generated in the simulator. In this paper, 
the driving policy $\policy_\theta$ is parameterized by a neural network which will output an action $a$ based on the state $s$ it observes. The simulator can generate driving behavior trajectories by rolling out $\policy_\theta$ for all vehicles simultaneously in the simulator. 
The optimization of the driving policy is optimized via TRPO~\cite{Schulman2015Trust} as in vanilla GAIL~\cite{ho2016generative}. 

\subsubsection{Downsampling of generated trajectories}
The goal of the downsampler is to construct the training data for interpolation, i.e., learning the mapping from a sparse trajectory to a dense one. For two consecutive points (i.e., $\traj^{t_s}$ and $\traj^{t_e}$ in generated sparse trajectory $\Traj_G$), we can sample a point $\traj^{t_i}$ in $\TrajD_G$ where $t_s\leq t_i\leq t_e$ and construct training samples for the interpolator.
The sampling strategies can be sampling at certain time intervals, sampling at specific locations or random sampling and we investigate the influence of different sampling rates in Section~\ref{sec:exp:sparsity}.

\subsubsection{Interpolation-Discriminator}
\begin{figure}[t!]
  \centering
  \includegraphics[width=.55\linewidth]{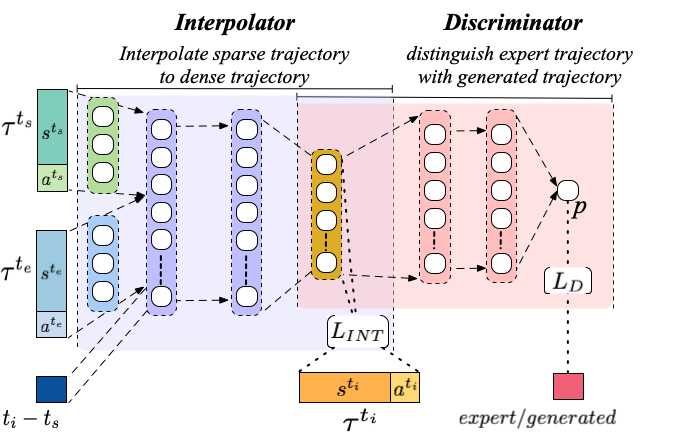}
  \caption{Proposed interpolation-discriminator network.}
  \label{fig:discr}
\end{figure}

The key difference between \learntosim and vanilla GAIL is in the discriminator. While learning to differentiate the expert trajectories from generated trajectories, the discriminator in \learntosim also learns to interpolate a sparse trajectory to a dense trajectory. Specifically, as is shown in Figure~\ref{fig:discr}, the proposed interpolation-discriminator copes with two subtasks in an end-to-end way: \emph{interpolation} on sparse data and \emph{discrimination} on dense data.

\paragraph{Interpolator module}
The goal of the interpolator is to interpolate the sparse expert trajectories $\Traj_E$ to the dense trajectories $\TrajD_E$. We can use the generated dense trajectories $\TrajD_G$ and sparse trajectories $\Traj_G$ from previous downsampling process as training data for the interpolator.  

For each point $\traj^{t_i}$ to be interpolated, we first concatenate state and action and embed them into an $m$-dimensional latent space:
\begin{equation}
h_s = \sigma(Concat(s^{t_s}, a^{t_s})W_s+b_s), h_e = \sigma(Concat(s^{t_e}, a^{t_e})W_e+b_e)
\end{equation}
where $K$ is the feature dimension after the concatenation of $s^{t_e}$ and $a^{t_e}$, $W_s\in \Real^{K\times M}$, $W_e\in \Real^{K\times M}$, $b_s\in \Real^{M}$ and $b_e\in \Real^{M}$ are weight matrix to learn, $\sigma$ is ReLU function (same denotation for the following $\sigma$). Here, considering $t_s$ and $t_e$ may have different effects on interpolation, we use two different embedding weights for $t_s$ and $t_e$.

After point embedding, we concatenate $h_s$ and $h_e$ with the time interval between $t_s$ and $t_i$, and use a multi-layer perception (MLP) with 
$L$ layers to learn the interpolation. 
\begin{equation}
    \centering
    \begin{aligned}
    h_{in} & = \sigma(Concat(h_s,h_e, t_i-t_s)W_0+b_0) \\
    h_1 & =  \sigma(h_{in}W_1+b_1), h_2 =  \sigma(h_{1}W_2+b_2), \cdots  \\
    h_L & = tanh(h_{L-1}W_{L}+b_{L}) = \hat{\traj}^{t_i} \\
    \end{aligned}
\end{equation}
where $W_0\in\Real^{(2M+1)\times N_0}$, $b_0\in\Real^{N_0}$ are the learnable weights; $W_j\in \Real^{N_j\times N_{j+1}}$ and $b_j\in \Real^{N_{j+1}}$ are the weight matrix for hidden layers ($1\leq j\leq L-1$) of interpolator; $W_L\in \Real^{N_j\times K}$ and $b_L\in \Real^{K}$ are the weight matrix for the last layer of interpolator, which outputs an interpolated point $\hat{\traj}^{t_i}$. In the last layer of interpolator, we use $tanh$ as all the feature value of $\traj^{t_i}$ is normalized to $[-1,1]$.

\paragraph{Discriminator module}
When sparse expert trajectories $\Traj_{E}$ are interpolated into dense trajectories $\TrajD_{E}$ by the interpolator, the discriminator module lears to differentiate between expert dense trajectories $\TrajD_{E}$ and generated dense trajectories $\TrajD_{D}$.
Specifically, the discriminator learns to output a high score when encountering an interpolated point $\hat{\traj}^{t_i}$ originated from $\TrajD_{E}$, and a low score when encountering a point from $\TrajD_{G}$ generated by $\policy_\theta$. The output of the discriminator $\DiscT(s, a)$ can then be used as a surrogate reward function whose value grows larger as actions sampled from $\policy_\theta$ look similar to those chosen by experts. 

The discriminator module is an MLP with $H$ hidden layers, takes $h_L$ as input and outputs the probability of the point belongs to $\Traj_E$.  
\begin{equation}
    \centering
    \begin{aligned}
    h^D_1 & =  \sigma(h_L W^D_1+b^D_1), h^D_2 =  \sigma(h^D_{1}W^D_2+b^D_2), \cdots  \\
   p & =  Sigmoid(h^D_{H-1}W^D_H+b^D_H) \\
    \end{aligned}
\end{equation}
where $W^D_i\in\Real^{N^D_{i-1}\times N^D_i}$, $b^D_i\in\Real^{N^D_i}$ are learnable weights for $i$-th layer in discriminator module. For $i=1$, we have $W^D_1\in\Real^{K\times N^D_1}$, $b^D_1\in\Real^{N^D_1}$, $K$ is the concatenated dimension of state and action; for $i=H$, we have $W^D_H\in\Real^{N^D_{H-1}\times 1}$, $b^D_H\in\Real$.

\paragraph{Loss function of Interpolation-Discriminator}
The loss function of the Interpolation-Discriminator network is a combination of interpolation loss $\Loss_{INT}$ and discrimination loss $\Loss_D$, which interpolates the unobserved points and predicts the probability of the point being generated by expert policy $\policy^E$ simultaneously, :
\begin{equation}
\label{eq:loss}
    \centering
    \begin{aligned}
    \Loss & = \lambda \Loss_{INT}+ (1-\lambda)\Loss_{D} = \lambda \mathbb{E}_{\traj^t\sim\traj\in\TrajD_G}(\hat{\traj}^t-\traj^t) + \\
       &   (1-\lambda)[\mathbb{E}_{\traj^{t}\sim\traj\in\Traj_G}\log p(\traj^t) + \mathbb{E}_{\traj^{t}\sim\traj\in\Traj_E}\log(1-p(\traj^t))]
    \end{aligned}
\end{equation}
where $\lambda$ is a hyper-parameter to balance the influence of interpolation and discrimination, $\hat{\traj}^{t}$ is the output of the interpolator module, and $p(\traj)$ is the output probability from the discriminator module.

\begin{algorithm}[bht]
\DontPrintSemicolon
\caption{Training procedure of \learntosim}
\label{alg:learntosim}

\KwIn{Sparse expert trajectories $\Traj_E$, initial policy and interpolation-discriminator parameters $\theta_0$, $\psi_0$}
\KwOut{Policy $\policy_\theta$, interpolation-discriminator $\indnet_\psi$}
\For{i $\longleftarrow$ 0, 1, \dots}
{
    Rollout dense trajectories for all agents $\TrajD_G = \{\traj |\traj = ( \traj^{t_0},\cdots ,\traj^{t_N} ),\ \traj^{t_j}=(s^{t_j}, a^{t_j})\sim \policy_{\theta_i} \}$; \;
    
    (Generator update step) \;
    
    $\bullet$ Score $\traj^{t_j}$ from $\TrajD_G$  with discriminator, generating reward using Eq.~\ref{eq:gail-reward}; \;
    
    $\bullet$ Update $\theta$ in generator given $\TrajD_G$ by optimizing Eq.~\ref{eq:gail}; \;
    
    (Interpolator-discriminator update step) \;
    
    $\bullet$ Interpolate $\Traj_E$ with the interpolation module in $\indnet$, generating dense expert trajectories $\TrajD_E$;  \;
    
    $\bullet$ Downsample generated dense trajectories $\TrajD_G$ to sparse trajectories $\Traj_G$; \;
    
    $\bullet$ Construct training samples for $\indnet$ \;
    
    $\bullet$ Update $\indnet$ parameters $\psi$ by optimizing Eq.~\ref{eq:loss} \;
}
\end{algorithm}

\subsection{Training and Implementation}
Algorithm~\ref{alg:learntosim} describes the \learntosim approach. 
In this paper, the driving policy is parameterized with a two-layer fully connected network with 32 units for all the hidden layers. The policy network takes the driving state $s$ as input and outputs the distribution parameters for a Beta distribution, and the action $a$ will be sampled from this distribution. The optimization of the driving policy is optimized via TRPO~\cite{Schulman2015Trust}. Following~\cite{bhattacharyya2018multi,kuefler2017imitating}, we use the features in Table~\ref{tab:features} to represent the driving state of a vehicle, and the driving policy takes the drivings state as input and outputs an action $a$ (i.e., next step speed). For the interpolation-discriminator network, each driving point is embedded to a 10-dimensional latent space, the interpolator module uses a three-layer fully connected layer to interpolate the trajectory and the discriminator module contains a two-layer fully connected layer. Some of the important hyperparameters are listed in Table~\ref{tab:parameters}.

\begin{table}[t!]
\centering
\caption{Features for a driving state}
\label{tab:features}
\begin{tabular}{cc}
\toprule
{Feature Type} & {Detail Features}\\
\midrule
Road network  & Lane ID, length of current lane, speed limit\\

Traffic signal & Current phase of traffic signal\\

Ego vehicle & \begin{tabular}[c]{@{}l@{}} Velocity, position in current lane,
distance to the next traffic signal\end{tabular} \\

Leading vehicle &  \begin{tabular}[c]{@{}l@{}} Relative distance, velocity and position in the current lane\end{tabular}\\

Indicators & \begin{tabular}[c]{@{}l@{}} Leading in current lane, exiting from intersection\end{tabular} \\
\bottomrule
\end{tabular}
\end{table}

\begin{table}[ht]
\centering
\caption{Hyper-parameter settings for \learntosim}
\label{tab:parameters}

\begin{tabular}{cc | cc}
\toprule
Parameter   & Value & Parameter   & Value \\ \midrule
\begin{tabular}[c]{@{}c@{}}Batch size for  generator\end{tabular}                   & 64 &
Batch size for \indnet                                                              & 32     \\
\begin{tabular}[c]{@{}c@{}}Update epoches for generator \end{tabular}   & 5   &
\begin{tabular}[c]{@{}c@{}}Update epoches for \indnet \end{tabular}    & 10     \\
\begin{tabular}[c]{@{}c@{}}Learning rate for  generator\end{tabular}                & 0.001 &
\begin{tabular}[c]{@{}c@{}}Learning rate for \indnet\end{tabular}                & 0.0001 \\
\begin{tabular}[c]{@{}c@{}}Number of layers in generator\end{tabular}               & 4  &
Balancing factor $\lambda$        & 0.5   \\     

\bottomrule
\end{tabular}
\end{table}

\section{Experiment}
\label{sec:experiment}

\subsection{Experimental Settings}
We conduct experiments on CityFlow~\cite{zhang2019cityflow}, an open-source traffic simulator that supports large-scale vehicle movements. In a traffic dataset, each vehicle is described as $(o, t, d, r)$, where $o$ is the origin location, $t$ is  time, $d$ is the destination location and $r$ is its route from $o$ to $d$. Locations $o$ and $d$ are both locations on the road network, and $r$ is a sequence of road ID. After the traffic data is fed into the simulator, a vehicle moves towards its destination based on its route. The simulator provides the state to the vehicle control method and executes the vehicle acceleration/deceleration actions from the control method.

\subsubsection{Dataset}
\begin{figure}[htbp]
  \centering
  \includegraphics[width=.95\linewidth]{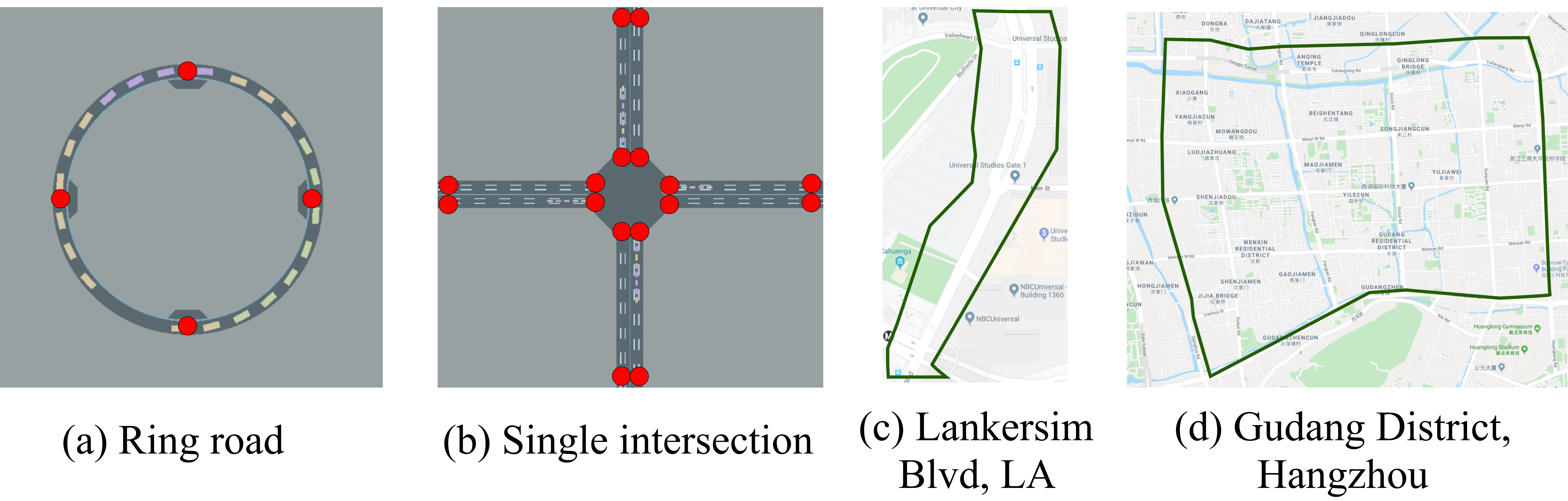}
  \caption{Illustration of road networks. (a) and (b) are synthetic road networks, while (c) and (d) are real-world road networks.}
  \label{fig:roadnet}
\end{figure}

In experiment, we use both synthetic data and real-world data. 
\paragraph{Synthetic Data}
In the experiment, we use two kinds of synthetic data, i.e., traffic movements under ring road network and intersection road network, as shown in Figure~\ref{fig:roadnet}. Based on the traffic data, we use default simulation settings of the simulator to generate dense expert trajectories and sample sparse expert trajectories when vehicles pass through the red dots.
~\noindent\\$\bullet$
\ringroad: The ring road network consists of a circular lane with a specified length, similar to~\cite{sugiyama2008traffic,wu2017emergent}. This is a very ideal and simplified scenario where the driving behavior can be measured. 
~\noindent\\$\bullet$
\intersectionone: A single intersection network with bi-directional traffic. The intersection has four directions (West$\rightarrow$East, East$\rightarrow$West, South$\rightarrow$North, and North$\rightarrow$South), and 3 lanes (300 meters in length and 3 meters in width) for each direction. Vehicles come uniformly with 300 vehicles/lane/hour in West$\leftrightarrow$East direction and 90 vehicles/lane/hour in South$\leftrightarrow$North direction.

\paragraph{Real-world Data}
We also use real-world traffic data from two cities: Hangzhou and Los Angeles. Their road networks are imported from OpenStreetMap\footnote{https://www.openstreetmap.org}, as shown in Figure~\ref{fig:roadnet}. The detailed descriptions of how we preprocess these datasets are as follows:
~\noindent\\$\bullet$
\LAfour. This is a public traffic dataset collected from Lankershim Boulevard, Los Angeles on June 16, 2005.
It covers an 1 $\times$ 4 arterial with four successive intersections. This dataset records the position and speed of every vehicle at every 0.1 second. We treat these records as dense expert trajectories and sample vehicles' states and actions when they pass through intersections as sparse expert trajectories.
~\noindent\\$\bullet$
\hangzhoufour. This dataset covers a 4 $\times$ 4 network of Gudang area in Hangzhou, collected from surveillance
cameras near intersections in 2016. This
region has relatively dense surveillance cameras and we sampled the sparse expert trajectories in a similar way as in \LAfour.

\subsubsection{Data Preprocessing}
To mimic the real-world situation where the roadside surveillance cameras capture the driving behavior of vehicles at certain locations, the original dense expert trajectories are processed to sparse trajectories by sampling the driving points near several fixed locations unless specified. We use the sparse trajectories as expert demonstrations for training models.
To test the imitation effectiveness, we use the same sampling method as the expert data and then compare the sparse generated data with sparse expert data. To test the interpolation effectiveness, we directly compare the dense generated data with dense expert data.

\begin{table}[t!!]
\centering
\caption{Statistics of dense and sparse expert trajectory in different datasets}
\label{tab:data}
\begin{tabular}{ccccc}
\toprule
   Env-name     & \ringroad & \intersectionone & \LAfour & \hangzhoufour \\ \midrule
Duration (seconds)   &   300  &  300   &  300  & 300 \\
\# of vehicles   &   22  &  109   &  321  & 425 \\
\begin{tabular}[c]{@{}c@{}} \# of points  (dense)\end{tabular}& 1996   & 10960 & 23009 & 87739\\
\begin{tabular}[c]{@{}c@{}}\# of points  (sparse)\end{tabular} & 40 & 283  & 1014  & 1481  \\
\bottomrule
\end{tabular}
\end{table}

\subsection{Compared Methods}
We compare our model with the following two categories of methods: calibration-based methods and imitation learning-based methods.

\paragraph{Calibration-based methods}
For calibration-based methods, we use Krauss model~\cite{krauss1998microscopic}, the default car-following model (CFM) of simulator SUMO \cite{SUMO2012} and CityFlow~\cite{zhang2019cityflow}. 
Krauss model has the following forms:
\begin{align}
v_{safe}(t) = v_l(t) + \frac{g(t)-v_l(t)t_r}{\frac{v_l(t) + v_f(t)}{2b}+t_r}\\
v_{des}(t) = \min [v_{safe}(t), v(t)+a\Delta t, v_{max}]
\end{align}
where $v_{safe}(t)$ the safe speed at time $t$, $v_l(t)$ and $v_f(t)$ is the speed of the leading vehicle and following vehicle respectively at time $t$, $g(t)$ is the gap to the leading vehicle, $b$ is the maximum deceleration of the vehicle and $t_r$ is the driver's reaction time. $v_{des}(t)$ is the desired speed, which is given by the minimum of safe speed, maximum allowed speed, and the speed after accelerating at $a$ for $\Delta t$. Here, $a$ is the maximum acceleration and $\Delta t$ is the simulation time step. 

We calibrate three parameters in Krauss model, which are the maximum deceleration of the vehicle, the maximum acceleration of the vehicle, and the maximum allowed speed.
\noindent\\$\bullet$
\textbf{Random Search (\cfmrs)}~\cite{asamer2013calibrating}: The parameters are chosen when they generate the most similar trajectories to expert demonstrations after a finite number of trial of random selecting parameters for Krauss model.
~\noindent\\$\bullet$
\textbf{Tabu Search (\cfmts)}~\cite{osorio2019efficient}: Tabu search chooses the neighbors of the current set of parameters for each trial. If the new CFM generates better trajectories, this set of parameters is kept in the Tabu list. 

\paragraph{Imitation learning-based methods}
We also compare with several imitation learning-based methods, including both traditional and state-of-the-art methods.
~\noindent\\$\bullet$
\textbf{Behavioral Cloning (\bc)} ~\cite{torabi2018behavioral} is a traditional imitation learning method. It directly learns the state-action mapping in a supervised manner.
~\noindent\\$\bullet$
\textbf{Generative Adversarial Imitation Learning (\gail)} is a GAN-like framework~\cite{ho2016generative}, with a generator controlling the policy of the agent, and a discriminator containing a classifier for the agent indicating how far the generated state sequences are from that of the demonstrations.

\subsection{Evaluation Metrics}

Following existing studies~\cite{kuefler2017imitating,bhattacharyya2018multi,zheng2020learing2sim}, to measure the error between learned policy against expert policy, we measure the position and the travel time of vehicles between generated dense trajectories and expert dense trajectories, which are defined as:
\begin{equation}
    RMSE_{pos} = \frac{1}{T}\mathop{\sum}_{t=1}^{T}\sqrt{\frac{1}{M}\mathop{\sum}_{i=1}^{m}(l_i^{t}-\hat{l}_i^{t})^2}, 
RMSE_{time} = \sqrt{\frac{1}{M}\mathop{\sum}_{i=1}^{m}(d_i-\hat{d}_i)^2}
\end{equation}
where $T$ is the total simulation time, $M$ is the total number of vehicles, $l_i^t$ and $\hat{l}_i^t$ are the position of $i$-th vehicle at time $t$ in the expert trajectories and in the generated trajectories relatively, $d_i$ and $\hat{d}_i$ are the travel time of vehicle $i$ in expert trajectories and generated trajectories respectively.

\begin{table*}[t!]
\centering
\caption{Performance w.r.t Relative Mean Squared Error (RMSE) of time (in seconds) and position (in kilometers). All the measurements are conducted on dense trajectories. Lower the better. Our proposed method \learntosim achieves the best performance.}\label{tab:overall_performance:interpolation}
\begin{tabular}{c|cc|cc|cc|cc}
\toprule
       & \multicolumn{2}{c|}{\ringroad}  & \multicolumn{2}{c|}{\intersectionone} & \multicolumn{2}{c|}{\LAfour}  & \multicolumn{2}{c}{\hangzhoufour}  \\ \cline{2-9}
       & time (s)  & pos (km) & time (s)  & pos (km) & time (s)  & pos (km)  & time (s)  & pos (km)     \\ \midrule
\cfmrs & 343.506 & 0.028  & 39.750 & 0.144  & 34.617 & 0.593 & 27.142  & 0.318\\
\cfmts & 376.593 & 0.025  & 95.330 & 0.184  & 33.298 & 0.510 & 175.326 & 0.359 \\
\bc    & 201.273 & 0.020  & 58.580 & 0.342  & 55.251 & 0.698 & 148.629 & 0.297 \\
\gail  & 42.061  & 0.023  & 14.405 & 0.032  & 30.475 & 0.445 & 14.973  & 0.196\\ \hline
\learntosim &  \textbf{16.970} & \textbf{0.018} & \textbf{4.550} & \textbf{0.024} & \textbf{19.671} & \textbf{0.405} & \textbf{5.254}   & \textbf{0.130}
\\\bottomrule     
\end{tabular}
\end{table*}

\subsection{Performance Comparison}
In this section, we compare the dense trajectories generated by different methods with the expert dense trajectories, to see how similar they are to the expert policy. The closer the generated trajectories are to the expert trajectories, the more similar the learned policy is to the expert policy.
From Table~\ref{tab:overall_performance:interpolation}, we can see that \learntosim achieves consistently outperforms over all other baselines across synthetic and real-world data. \cfmrs and \cfmrs can hardly achieve satisfactory results because the model predefined by CFM could be different from the real world. Specifically, \learntosim outperforms vanilla \gail, since \learntosim interpolates the sparse trajectories and thus has more expert trajectory data, which will help the discriminator make more precise estimations to correct the learning of policy.

\subsection{Study of \learntosim}
\paragraph{Interpolation Study}
\label{sec:exp:sparsity}
 
To better understand how interpolation helps in simulation, we compare two representative baselines with their two-step variants. Firstly, we use a pre-trained non-linear interpolation model to interpolate the sparse expert trajectories following the idea of~\cite{yi2016st,tang2019joint}. Then we train the baselines on the interpolated trajectories.

Table~\ref{tab:interpolation_performance} shows the performance of baseline methods in \ringroad and \intersectionone. We find out that baseline methods in a two-step way show inferior performance. One possible reason is that the interpolated trajectories generated by the pre-trained model could be far from the real expert trajectories when interacting in the simulator. Consequently, the learned policy trained on such interpolated trajectories makes further errors. 

In contrast, \learntosim learns to interpolate and imitate the sparse expert trajectories in one step, combining the interpolator loss and discriminator loss, which can propagate across the whole framework. If the trajectories generated by $\policy_\theta$ is far from expert observations in current iteration, both the discriminator and the interpolator will learn to correct themselves and provide more precise reward for learning $\policy_\theta$ in the next iteration. Similar results can also be found in \LAfour and \hangzhoufour, and we omit these results due to page limits.

\begin{table}[t!]
\centering
\caption{RMSE on time and position of our proposed method \learntosim against baseline methods and their corresponding two-step variants. Baseline methods and \learntosim learn from sparse trajectories, while the two-step variants interpolate sparse trajectories first and trained on the interpolated data. \learntosim achieves the best performance in most cases.}\label{tab:interpolation_performance}
\begin{tabular}{c|cc|cc}
\toprule
       & \multicolumn{2}{c|}{\ringroad}                                                    & \multicolumn{2}{c}{\intersectionone}                                                 \\ 
       & time (s)        & position (km)       & time (s)        & position (km)  \\ \midrule
\cfmrs   & 343.506  & 0.028    & 39.750  & 0.144  \\
\begin{tabular}[c]{@{}c@{}}\cfmrs (two step) \end{tabular}& 343.523 & 0.074  & 73.791  & 0.223  \\ \hline
\gail   & 42.061   & 0.023    & 14.405    & 0.032  \\
\begin{tabular}[c]{@{}c@{}}\gail (two step) \end{tabular}  & 98.184  & 0.025 &  173.538 & 0.499  \\ \hline
\learntosim  &  \textbf{16.970}  & \textbf{0.018}  & \textbf{4.550}    & \textbf{0.024}            
\\\bottomrule     
\end{tabular}
\end{table}

\begin{figure}[ht!]
  \centering
  \begin{tabular}{cccc}
  \includegraphics[width=0.23\textwidth]{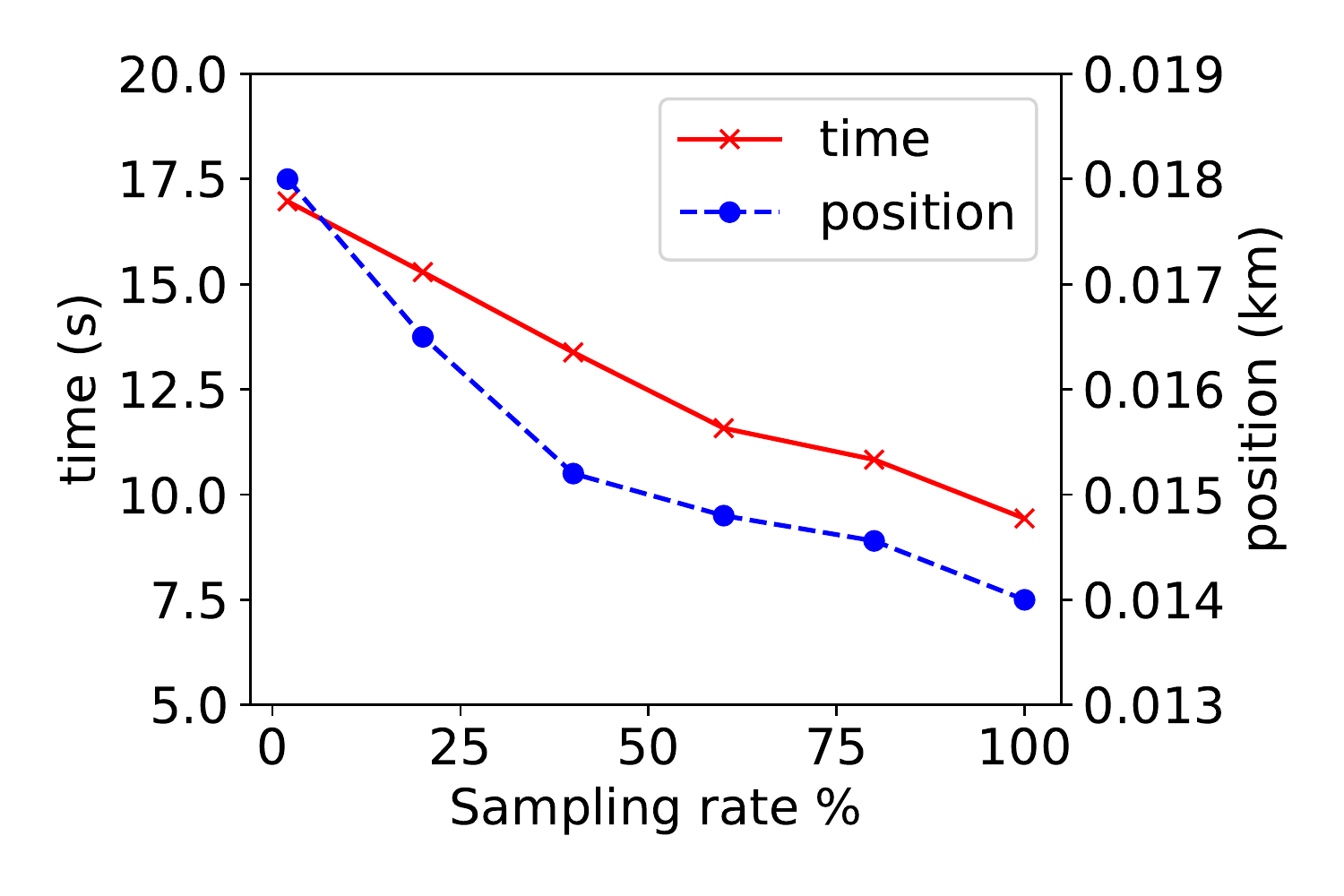}&
  \includegraphics[width=0.23\textwidth]{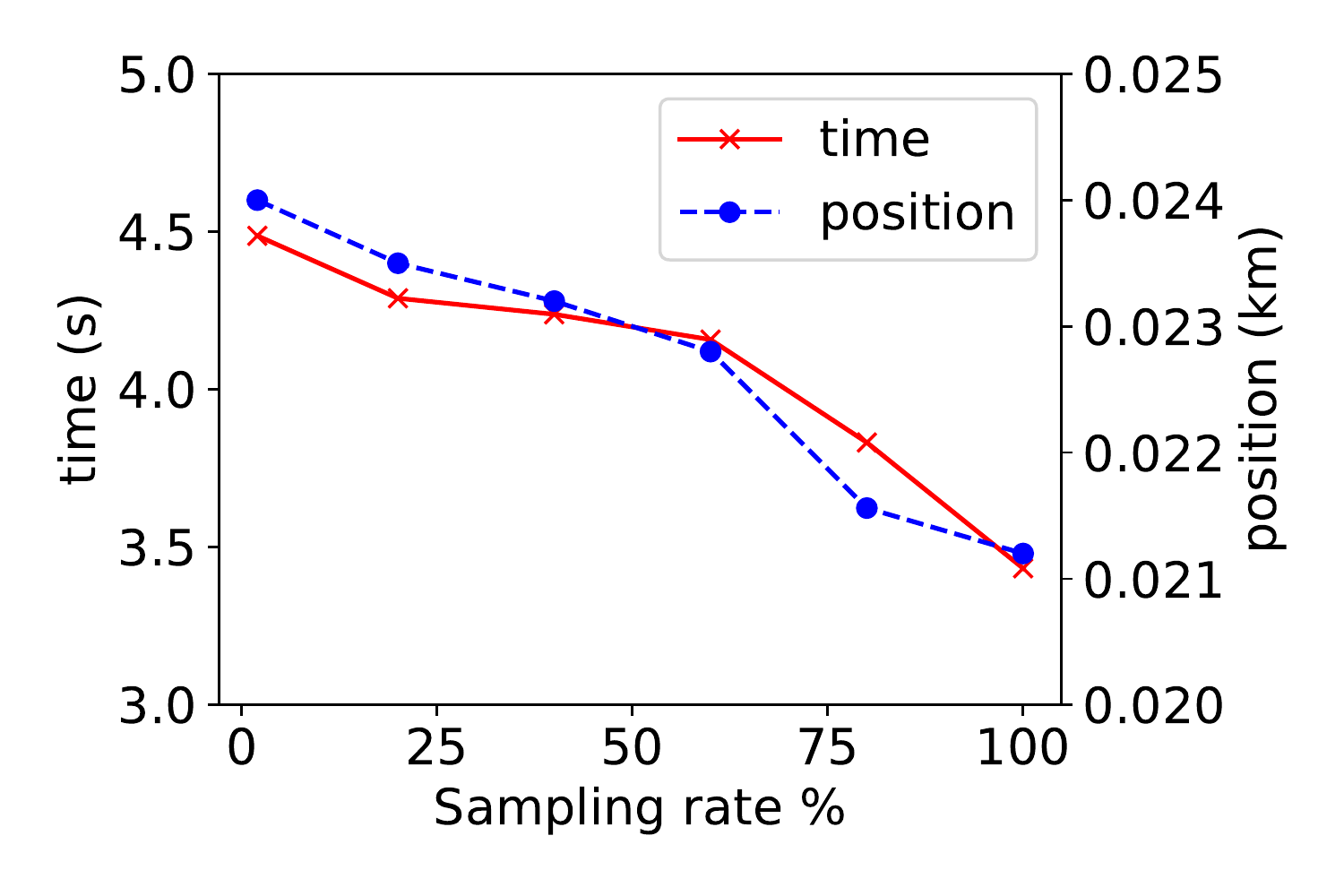} &
  \includegraphics[width=0.23\textwidth]{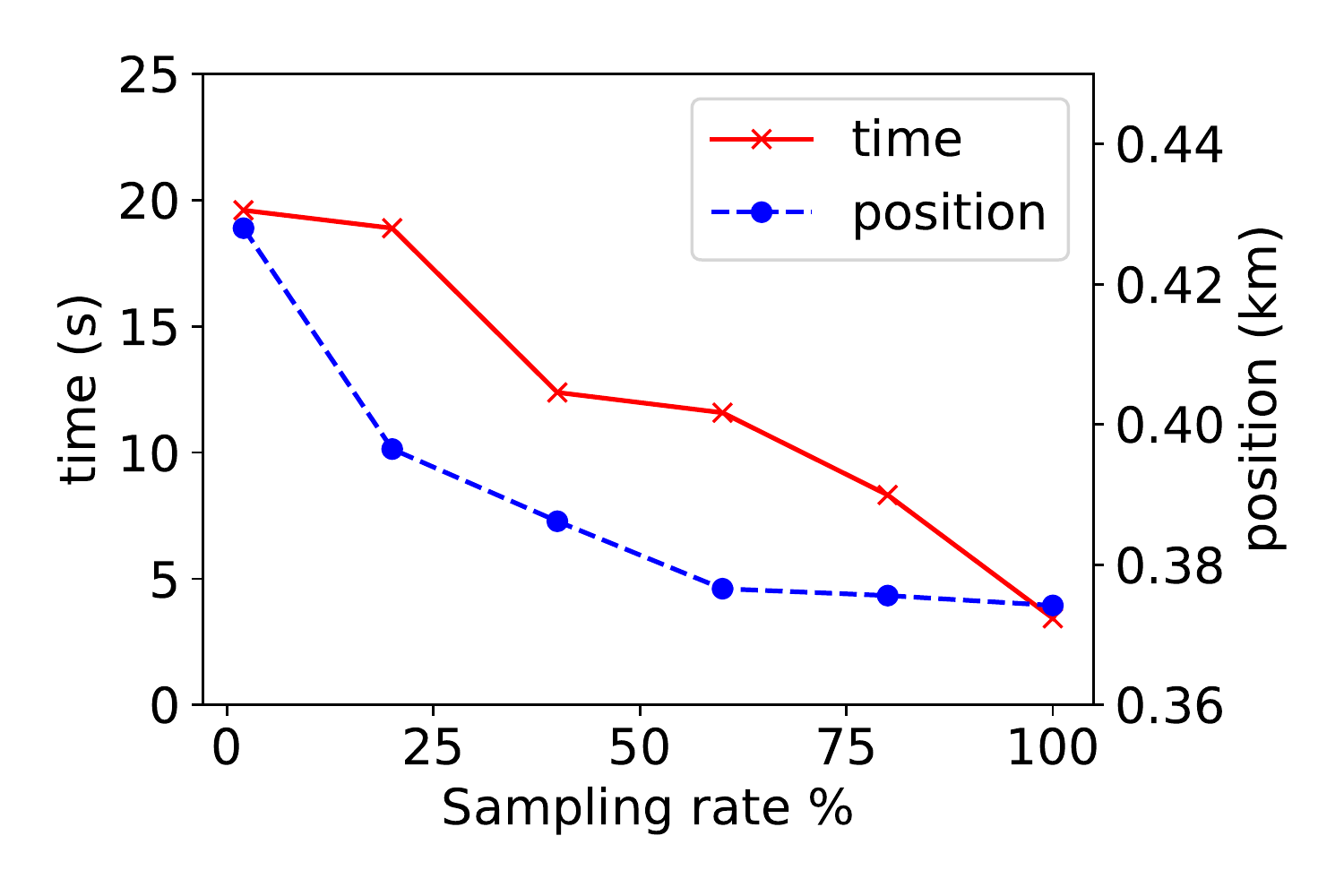}&
  \includegraphics[width=0.23\textwidth]{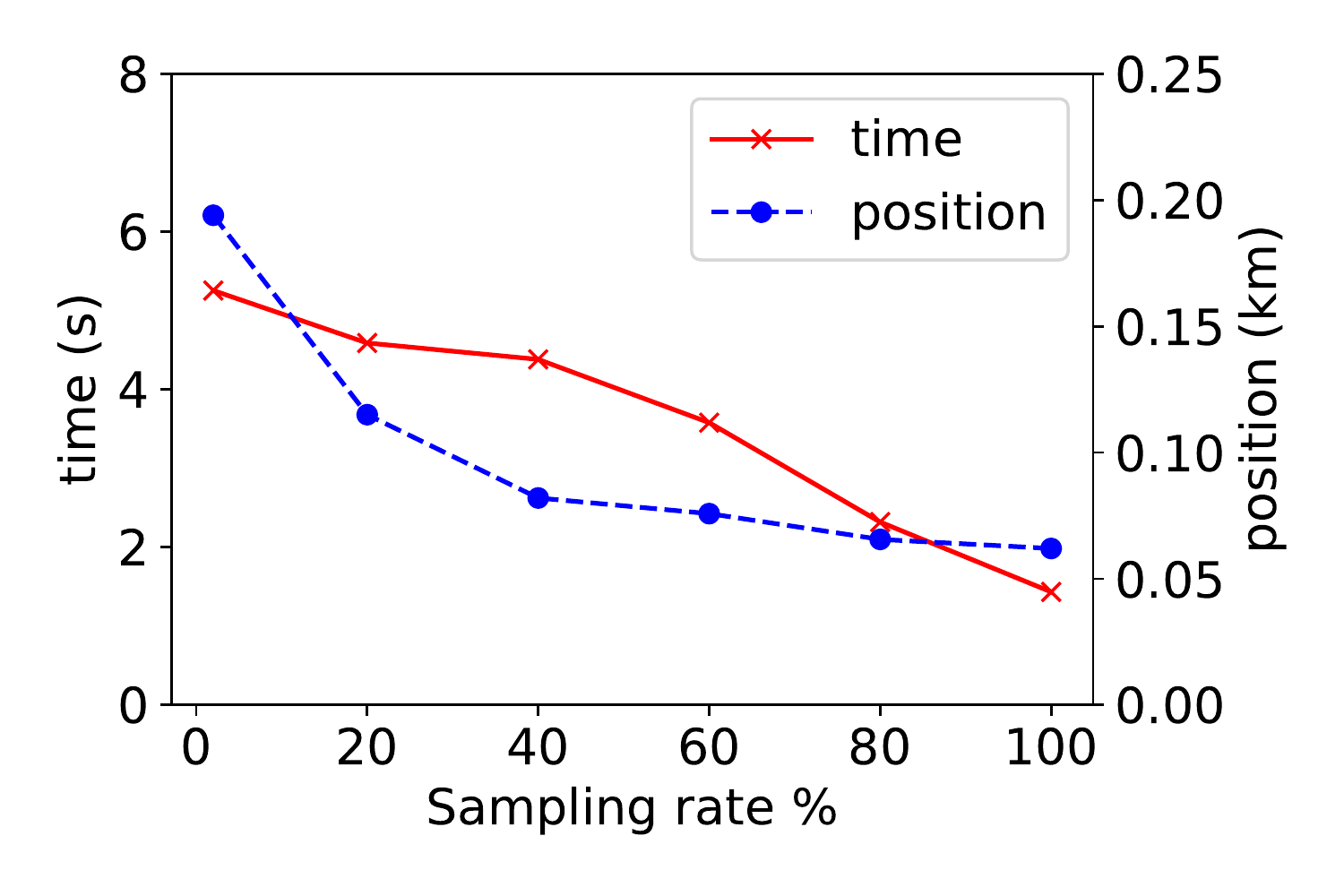}\\
   (a) \ringroad & (b) \intersectionone  & (c) \LAfour & (d) \hangzhoufour  \\
  \end{tabular}
  \caption{RMSE on time and position of our proposed method \learntosim under different level of sparsity. As the expert trajectory become denser, a more similar policy to the expert policy is learned.}
  \label{fig:sparsity-study}
\end{figure}

\paragraph{Sparsity Study}
In this section, we investigate how different sampling strategies influence \learntosim. We sample randomly from the dense expert trajectories at different time intervals to get different sampling rates: 2\%, 20\%, 40\%, 60\%, 80\%, and 100\%. We set the sampled data as the expert trajectories and evaluate by measuring the performance of our model in imitating the expert policy. As is shown in Figure~\ref{fig:sparsity-study}, with denser expert trajectory, the error of \learntosim decreases, indicating a better policy imitated by our method.

\subsection{Case Study}
To study the capability of our proposed method in recovering the dense trajectories of vehicles, we showcase the movement of a vehicle in \ringroad data learned by different methods. 

\begin{figure}[t!]
  \centering
  \includegraphics[width=.95\linewidth]{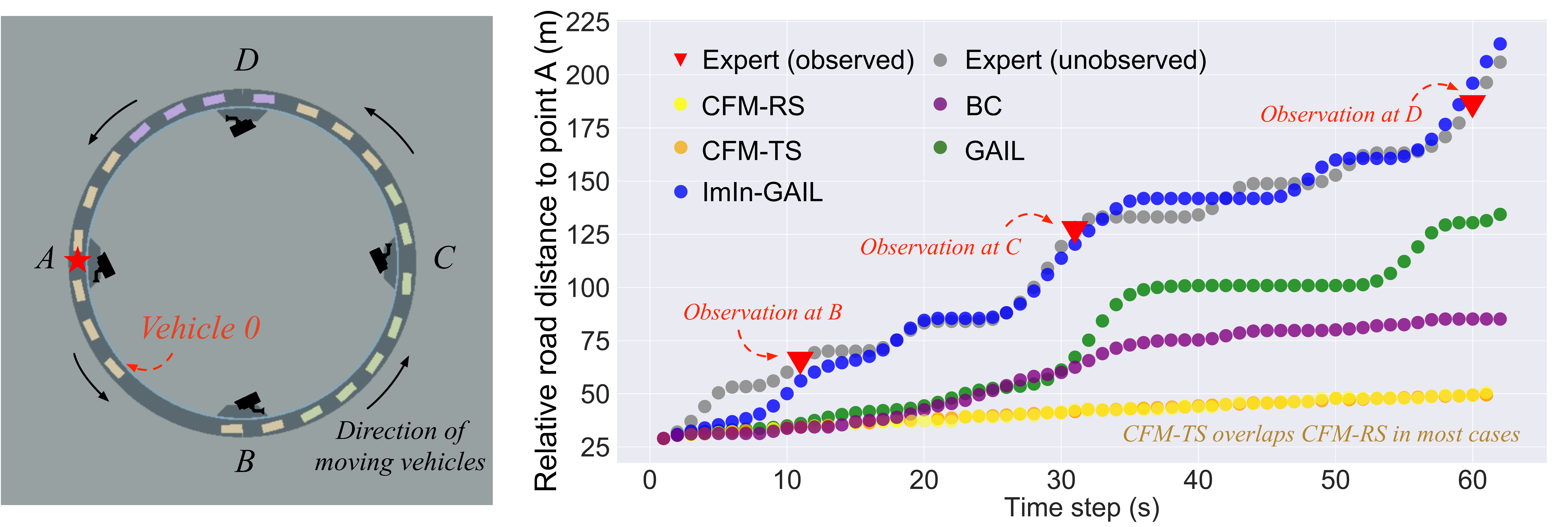}
  \caption{The generated trajectory of a vehicle in the \ringroad scenario. Left: the initial position of the vehicles. Vehicles can only be observed when they pass four locations $A$, $B$, $C$ and $D$ where cameras are installed.  Right: the visualization for the trajectory of $Vehicle\ 0$. The x-axis is the timesteps in seconds. The y-axis is the relative road distance in meters. Although vehicle 0 is only observed three times (red triangles), \learntosim (blue points) can imitate the position of the expert trajectory (grey points) more accurately than all other baselines. Better viewed in color.}
  \label{fig:case_study}
\end{figure}

We visualize the trajectories generated by the policies learned with different methods in Figure~\ref{fig:case_study}. We find that imitation learning methods (\bc, \gail, and \learntosim) perform better than calibration-based methods (\cfmrs and \cfmts). This is because the calibration based methods pre-assumes an existing model, which could be far from the real behavior model. On the contrast, imitation learning methods directly learn the policy without making unrealistic formulations of the CFM model. Specifically, \learntosim can imitate the position of the expert trajectory more accurately than all other baseline methods. The reason behind the improvement of \learntosim against other methods is that in \learntosim, policy learning and interpolation can enhance each other and result in significantly better results.

\section{Related Work}
\label{sec:related}

\paragraph{Parameter calibration} In parameter calibration-based methods, the driving behavior model is a prerequisite, and parameters in the model are tuned to minimize a pre-defined cost function. Heuristic search algorithms such as random search, Tabu search\cite{osorio2019efficient}, and genetic algorithm \cite{kesting2008calibrating} can be used to search the parameters. These methods rely on the pre-defined models (mostly equations) and usually fail to match the dynamic vehicle driving pattern in the real-world.

\paragraph{Imitation learning} Without assuming an underlying physical model, we can solve this problem via imitation learning. There are two main lines of work: (1) behavior cloning (BC) and Inverse reinforcement learning (IRL). BC learns the mapping from demonstrated observations to actions in a supervised learning way~\cite{michie1990cognitive,torabi2018behavioral}, but suffers from the errors which are generated from unobserved states during the simulation. On the contrast, IRL not only imitates observed states but also learns the expert's underlying reward function, which is more robust to the errors from unobserved states~\cite{abbeel2004apprenticeship,ng2000algorithms,ziebart2008maximum}. Recently, a more effective IRL approach, GAIL~\cite{ho2016generative}, incorporates generative adversarial networks with learning the reward function of the agent. However, all of the current work did not address the challenges of sparse trajectories, mainly because in their application contexts, e.g., game or robotic control, observations can be fully recorded every time step.


\section{Conclusion}

In this paper, we present a novel framework \learntosim to integrate interpolation with imitation learning and learn the driving behavior from sparse trajectory data. Specifically, different from existing literature which treats data interpolation as a separate and preprocessing step, our framework learns to interpolate and imitate expert policy in a fully end-to-end manner. Our experiment results show that our approach significantly outperforms state-of-the-art methods. The application of our proposed method can be used to build a more realistic traffic simulator using real-world data.

\label{sec:conclusion}

\section*{Acknowledgments}
The work was supported in part by NSF awards \#1652525 and \#1618448. The views and conclusions contained in this paper are those of the authors and should not be interpreted as representing any funding agencies.

\bibliographystyle{splncs04}
\bibliography{ecmlpkdd20}

\end{document}